# Back-of-the-Book Index Automation for Arabic Documents

**Nawal Haidar** and **Fadi A. Zaraket**
Arab Center for Research and Policy Studies, Doha
`{nhaidar,fzaraket}@dohainstitute.edu.qa`

## Abstract

Back-of-the-book indexes are crucial for book readability. Their manual creation is laborious and error prone. In this paper, we consider automating back-of-the-book index extraction for Arabic books to help simplify both the creation and review tasks. Given a back-of-the-book index, we aim to check and identify the accurate occurrences of index terms relative to the associated pages. To achieve this, we first define a pool of candidates for each term by extracting all possible noun phrases from paragraphs appearing on the relevant index pages. These noun phrases, identified through part-of-speech analysis, are stored in a vector database for efficient retrieval. We use several metrics, including exact matches, lexical similarity, and semantic similarity, to determine the most appropriate occurrence. The candidate with the highest score based on these metrics is chosen as the occurrence of the term. We fine-tuned a heuristic method, that considers the above metrics and that achieves an F1-score of .966 (precision=.966, recall=.966). These excellent results open the door for future work related to automation of back-of-the-book index generation and checking.

## 1 Introduction

In non-fiction literature, the inclusion of a back-of-the-book index (BoBI) is standard practice. BoBI provides readers with a valuable tool for quickly identifying important terminology and locating specific information within the text. However, manually constructing and reviewing BoBI is a time-consuming task prone to errors. Automating the extraction and reviewing of existing BoBIs is thus essential, enabling efficient navigation and error detection.

We distinguish between two tools: ArBoBIM, the Arabic BoBI mapper which extracts an existing BoBI and maps its index terms to their occurrences, and ArBoBIE, the Arabic BoBI extractor which generates a BoBI for Arabic documents. This paper reports on our efforts to develop necessary resources and build ArBoBIM.

To build the resources, we transform existing raw text BoBIs, typically found at the end of a book, into accurate structured and programmable index maps. These maps provide detailed descriptions of the occurrences of each term, including exact form and position within text. BoBIs already contain page numbers for each term, however several challenges exist.

First, to help editors and reviewers, the task involves aligning the BoBI page numbers with the corresponding pages within the container working document. This is usually provided in Microsoft Word (.docx) documents.

This presents a challenge because the page numbers in the index are often based on production versions of the books. These include final layout and style elements added at the last stages by graphic designers and the layout management team. Those typically perform the task with professional high quality graphic tools such as Adobe Illustrator and produce their output in a production portable document format (.pdf). This changes several elements in the books, in particular pagination and requires changes in the BoBI among other tasks.

Thus, often times, BoBI editors defer inclusion of page numbers in the editing stage to avoid duplicate work in the last stage.

Second, we need to identify the top match(es) for each term within the defined search scope, ensuring accuracy and relevance. Notice that more than one candidate may exist in the same page.

To address these challenges, we developed and used existing robust algorithms and natural language processing (NLP) tools and techniques. These include text extraction, page layout analysis, pattern matching, named entity recognition (NER), morphological analysis, and lexical and semantic similarity measures.

By providing detailed descriptions of term occurrences and leveraging NLP techniques, ArBoBIM enhances navigation and review processes for Arabic documents.

We make the annotated dataset available online for the research community.

This paper will delve into the methodology and results of our work, demonstrating the effectiveness of our approach in overcoming the identified challenges and improving efficiency in BoBI automation.

## 2 Related Work

Research on BoBI extraction methods has evolved with the development of automated techniques. TMG-BoBI utilizes text mining and the Text-to-Matrix Generator to automate BoBI creation (Koutropoulou and Gallopoulos, 2019). It employs Automated Keyword Extraction (AKE) to identify relevant terms, including unigrams, bigrams, and trigrams, and integrates Part-of-Speech (PoS) tagging to categorize words by their linguistic roles, optimizing index readability with minimal user intervention. While effective with English and Latin text, its performance is not satisfactory for Arabic.

Although distinct, keyword extraction plays a crucial role in enhancing the efficiency and accuracy of back-of-the-book index creation by automating the identification of relevant terms and concepts within the text.

An Unsupervised framework for keyword extraction (Mao et al., 2020) focuses on candidate keyword selection and word scoring. They enhance keyword extraction by incorporating word co-occurrence and semantic relationships, resulting in improvements over baseline methods.

Arabic language text mining presents unique challenges due to its intricate grammar, rich morphology, and relatively rich semantic diversity. This makes direct application of keyword extraction techniques ineffective. The KpST system (Sahmoudi and Lachkar, 2016) leverages a Suffix Tree Data Structure for Arabic keyphrase extraction, employing linguistic patterns and an adapted C-value method to extract relevant keyphrases. By tackling the complexities of Arabic morphology, the KpST system improved keyphrase extraction for Arabic documents. While effective at creating terms that maybe useful for search engine optimization, the selected keywords are directly suitable for BoBI as those require additional relevance and presentation criteria.

In the light of this, ArBoBIM represents a tool useful for data collection and review processes of BoBIs generated manually or automatically.

## 3 Methodology

This section describes the data processing, page mapping, and entry mapping methodology we used to develop ArBoBIM.

### 3.1 Index Entry Data Processing

To transform the BoBI to a structured index map, we first need to extract it as raw text from books. Our source materials are obtained from the Arab Center for Research and Policy Studies (ACRPS), covering a range of topics including politics, sociology, and history, in both Word (.docx) and PDF formats. We start working with the Word version, as text extraction is more straightforward. We utilize the python-docx module to process the text and format of the documents.

ArBoBIM extracts the BoBI, which is typically located at the back of the book, as a raw text and then processes it to generate lists of index terms and their corresponding page numbers.

ArBoBIM considers special features including multiple points of entry (MPE), hierarchical structure, as well as style variations with specific phrases.

MPE signify the utilization of terms with similar meanings but differing terminology, such as "المملكة المتحدة ينظر بريطانيا" (United kingdom aka Britain). ArBoBIM stores MPEs separately for future reference.

We differentiate between two approaches with MPE:

- Terms are treated equally without differentiation, each sharing the same page numbers referenced under a single entry;

- Terms are distinguished, with each term having its own designated page numbers.

Hierarchical structure involves organizing terms into sub-entries under main entries to address specificity or subtopics. The following example illustrates the United Nations with the Security council as a sub-entry.

الأمم المتحدة
-- مجلس الأمن

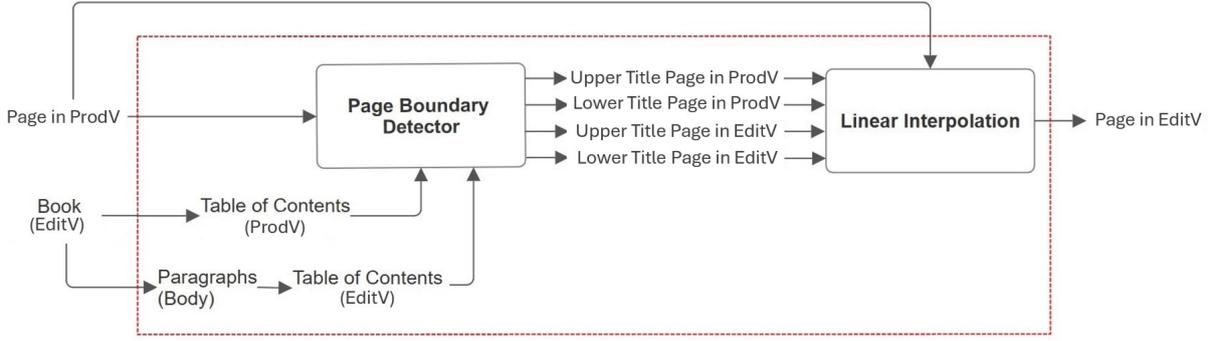

Figure 1: Production (ProdV) to Editing (EditV) Versions Page Mapper

In addition, certain cases involve phrase inversions, such as when the term is a name of a person. For example, "ناصر أبو رحمة" is indexed as "أبو رحمة، ناصر".

## 3.2 Production to Editing Page Mapping

The page numbers extracted from the raw BoBI correspond to the production version of the book in PDF format. This is because the PDF version includes the final formatting and style elements. The page numbers in the BoBI are specified manually with reference to the final version. To locate these terms in the editing version usually in Word format, we must establish a mapping between PDF and Word pages. Figure 1 illustrates the process.

Comparison between production and editing versions show that differences in page numbers stem from structural aspects, such as font size, line spacing, title styling, margins, and inclusion of graphics. Consequently, the mapping from production to editing is nearly linear.

To compute this mapping, ArBoBIM extracts the table of contents from the production version of the book.

Next, ArBoBIM traverses the content of the book, identifying titles and extracting their respective page numbers in the editing format. With the table of contents providing page references for titles and section and subsection headings in both production and editing versions, ArBoBIM determines a range of search for the production pages bounding a given production page within a range pages in the editing version.

ArBoBIM then employs linear interpolation to map a given production page to its corresponding editing page, with a window sliding by one to accommodate for potential errors.

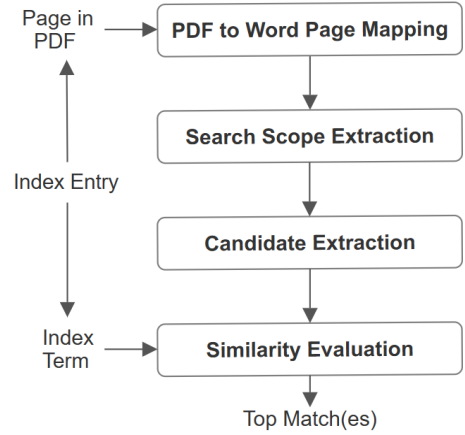

Figure 2: ArBoBIM Operation

## 3.3 Mapping Each Term to its Occurrences

ArBoBIM traverses the whole book and extracts its paragraphs with their corresponding details, including the page numbers and section titles. For each index term, ArBoBIM maps its pages to the editing version and consequently identifies the list of paragraphs belonging to these pages, which now constitute the search scope for a given term.

ArBoBIM initially searches for exact matches of the term within the search scope. If it finds an exact match, it records its position and returns the occurrence.

Otherwise, ArBoBIM looks for phrases having lexical similarity above a specified threshold with the index term. It uses morphological analysis through CAMeL Tools (Obeid et al., 2020) to obtain the lemmas of the words. Then, it uses the Levenshtein distance (Levenshtein, 1966) to compute a lexical similarity ratio.

Lastly, if no exact and lexical matches are found, ArBoBIM resorts to semantic similarity (Slimani, 2013).

ArBoBIM defines the pool of candidates as all possible noun phrases in the search scope, in which it uses Part-of-Speech (POS) tagging using CAMeL Tools (Obeid et al., 2020) and Named Entity Recognition (NER) using Wojood (Jarrar et al., 2022). POS tags specify the role of a word in a sentence such a noun, verb, adverb or particle. It is key to detect noun phrases in the search scope. An example follows.

| استخدمت | لجنة | تمثل | الأمم | المتحدة |
|---|---|---|---|---|
| verb | noun | verb | noun | adj |

NER identifies typed entities, such as names of people, organizations, locations, and dates, within a text. Wojood (Jarrar et al., 2022) identifies 21 entity classes in Arabic with high accuracy. Wojood returns organization (ORG) for the phrase 'الأمم المتحدة' (United Nations).

Using POS tagging, ArBoBIM defines a set of rules that a noun phrase should follow. For example, a noun phrase should start with a noun, may include an adjective, cannot include a verb, and may not end with a preposition. These rules work well for limiting the number of candidate phrases.

ArBoBIM stores the extracted noun phrases, along with their metadata including paragraph ID, in a vector database using ChromaDB (chr, 2023) and leverages the "distiluse-base-multilingual-cased-v1" model from the SentenceTransformer library (Reimers and Gurevych, 2019) for embeddings. Subsequently, it queries for candidates with top semantic similarity with the index term.

ArBoBIM then calculates the lexical similarity ratio with the index term for each of the top candidates, and returns an aggregation of both lexical and semantic similarities, as a score for the occurrence of the index term on the page in question. For example, the top candidates for the index term "العلاقات الدولية" (international relations) follow.

- إجماع دولي (international consensus)
- القانون الدولي (international law)
- الصحف الدولية (international newspapers)
- الدول الأجنبية (foreign countries)
- العلاقات على المستوى الدولي (relations on international level)
- الضغوط الدولية (international pressures)
- الإرهاب (terrorism)

"العلاقات على المستوى الدولي" scores highest .67 lexical and .92 semantic similarity.

ArBoBIM uses the measures above to decide on the top candidates relative to both the index term

Table 1: Evaluation Metrics

| Metric | Value |
|---|---|
| Precision | 0.966 |
| Recall | 0.966 |
| F1 Score | 0.966 |

and its other points of entries in the case where they all share the same list of pages.

## 4 Results

We applied a threshold of .8 for lexical similarity, which resulted in numerous false positives. Subsequently, we adjusted the threshold to .9, which significantly improved precision. This higher threshold also filtered out many correct occurrences reducing recall. Semantic similarity was able to correctly detect and recover them. Table 1 shows an F-score of .966 with similar precision and recall.

Upon inspection, errors arise from several factors. First, we detected inaccuracies in mapping pages from PDF to Word versions leading to incorrect search scopes. This can be remedied by increasing the sliding window at the expense of expanding the search scope. CAMel and Wojood inaccuracies also contributed to the omission of noun phrases that would have been true positives otherwise. This can also be remedied by loosening the rules for noun phrase inclusion. Finally, BoBI concepts sometimes refer to paragraphs discussing the concept without including the direct term in the paragraph. This requires models with reasoning and language understanding capacities that we plan to employ in the future.

## 5 Conclusion

In this paper, we have presented a method to extract the back-of-the-book index of Arabic documents by transforming it from raw text to a structured index map to locate the occurrences of index terms in their context. This is important to automate navigation and review processes for BoBIs, and to build resources for BoBI creation. ArBoBIM demonstrated excellent results and we identified opportunities for further improvements that we will seek in the near future.

## Ethics Statement